\documentclass[conference]{IEEEtran}
\IEEEoverridecommandlockouts
\usepackage{cite}
\usepackage{amsmath,amssymb,amsfonts}
\usepackage{algorithmic}
\usepackage{graphicx}
\usepackage{textcomp}
\usepackage{xcolor}
\usepackage[table]{xcolor}

\usepackage[T1]{fontenc}
\usepackage{times}
\usepackage{tabularx}
\usepackage{booktabs}
\usepackage{multirow} 
\usepackage[font=small,skip=2pt]{caption}
\usepackage{booktabs}   
\usepackage{array} 
\usepackage{amssymb}
\usepackage{etoolbox}
\usepackage{makecell}
\usepackage{diagbox} 
\def\BibTeX{{\rm B\kern-.05em{\sc i\kern-.025em b}\kern-.08em
    T\kern-.1667em\lower.7ex\hbox{E}\kern-.125emX}}
\begin{document}

\setlength{\abovedisplayskip}{2pt}
\setlength{\belowdisplayskip}{2pt}
\setlength{\abovedisplayshortskip}{1pt}
\setlength{\belowdisplayshortskip}{1pt}
\setlength{\textfloatsep}{6pt}
\setlength{\floatsep}{4pt}
\setlength{\intextsep}{6pt}

\title{An Edge-aware Prompt-enhanced SAM for Ultrasound Image Segmentation\\
\thanks{{*}Corresponding authors}}

\author{
\IEEEauthorblockN{Wenhao Li\textsuperscript{1}, Fangyi Liu\textsuperscript{2*}, Bo Du\textsuperscript{2*}}
\IEEEauthorblockA{\textsuperscript{1}\textit{Key Laboratory of Aerospace Information Security and Trusted Computing, Ministry of Education,} \\
\textit{School of Cyber Science and Engineering, Wuhan University, Wuhan, China}}
\IEEEauthorblockA{\textsuperscript{2}\textit{National Engineering Research Center for Multimedia Software, Institute of Artificial Intelligence,} \\
\textit{School of Computer Science and Hubei Key Laboratory of Multimedia and Network Communication Engineering,}\\
\textit{Wuhan University, Wuhan, China}}
\IEEEauthorblockA{2024202210100@whu.edu.cn, fangyiliu@whu.edu.cn, dubo@whu.edu.cn}
}

\maketitle

\begin{abstract}

Ultrasound image segmentation is essential for delineating anatomical structures and lesions, providing the foundation for accurate diagnosis. While the Segment Anything Model (SAM) has demonstrated remarkable success on natural images, its performance on ultrasound data is often hindered by poor boundary delineation. To address this limitation, we propose EP-SAM, an edge-aware and prompt-enhanced adaptation of SAM. Specifically, we leverage multi-block feature extraction from the image encoder to enrich coarse-to-fine semantic representations, while edge-aware supervision of the image encoder improves robustness to contour ambiguity and speckle noise. By integrating these complementary cues, EP-SAM generates high-quality prompts that effectively guide the model toward target regions of interest. Experimental results on multiple benchmarks demonstrate that EP-SAM consistently outperforms existing SAM-based methods.

\end{abstract}


\section{Introduction}

Ultrasound medical image segmentation is crucial for clinical diagnosis, treatment planning, and biomedical research~\cite{liu2021review}. Owing to its radiation-free, portable, and cost-effective nature, ultrasound is widely adopted for point-of-care applications. However, its real-time, reflection-based imaging mechanism introduces speckle noise, low contrast, and blurred anatomical boundaries, which degrade image quality and obscure structural details~\cite{zhang2025adapting}. These inherent challenges not only complicate accurate segmentation but also hinder the direct application of deep learning models and prompt-based strategies originally designed for natural images.

\begin{figure}[t]
\centering
\includegraphics[width=0.95\columnwidth]{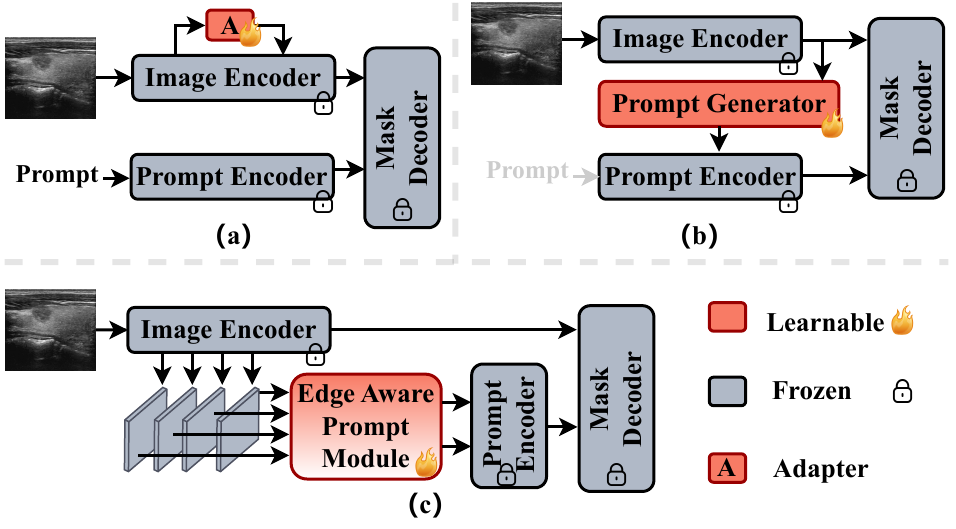} 
\caption{Comparison of SAM-based adaptations for medical image segmentation. Unlike other prompting strategies, our EP-SAM enhances the synergy of individual components and explicitly alleviates boundary ambiguity.}
\label{fig1}
\end{figure}

Recently, the Segment Anything Model (SAM) has demonstrated remarkable segmentation capabilities. This breakthrough has further opened new opportunities for advancing ultrasound image segmentation and has motivated a growing body of SAM-based methods, which can be broadly categorized as follows. (1) Decoder-level adaptation focuses on refining the segmentation head~\cite{ma2024segment} by fine-tuning the mask decoder to better align with medical domain data. (2) Encoder-level adaptation modifies the image encoder to enhance feature representation~\cite{zhang2023customized, wu2025medical}, and (3) Prompt-level adaptation optimizes or learns new prompts to guide the segmentation results without modifying SAM’s weights~\cite{yan2025pgp}. However, as shown in Figure~\ref{fig1}(a), most existing adaptations treat the constituent components of SAM as isolated modules, resulting in a lack of synergy between them. This architecture often fails to leverage the rich semantic information available during the encoding stage to guide the subsequent prompting process. To address this, self-prompting methods have emerged as a promising solution. As illustrated in Figure~\ref{fig1}(b), they generate the prompts required by the SAM prompt encoder by leveraging the final features from the image embedding.

Despite these advances, SAM~\cite{kirillov2023segment} relies heavily on high-level semantic cues and tends to overlook fine-grained anatomical boundaries in ultrasound images. As a result, it struggles to accurately delineate target edges, especially in regions with blurred organ boundaries or gradual tissue transitions, often leading to ``edge drift". 
To address this issue, we propose EP-SAM, as shown in Figure~\ref{fig1}(c), a structurally enhanced adaptation of the Segment Anything Model for ultrasound image segmentation, which consists of two main modules.

To explicitly preserve fine-grained boundary cues during the encoding process, we propose an Edge-Aware Module. This module consists of multiple carefully designed Edge-Aware Blocks, aimed at extracting meaningful ultrasound boundaries from the intermediate features of the SAM image encoder. Unlike traditional approaches that rely on heuristic operators like Sobel for supervision, each block within the EAM is supervised using ground-truth-derived boundaries, ensuring high-fidelity contour alignment. Architecturally, the EAM serves as a dynamic bridge between the image and prompt encoders. By explicitly injecting refined boundary cues into the prompt generation process, the model achieves boundary-informed prompt evolution.

To optimize the synergy between the image and prompt encoders, we introduce the Prompt Enhanced Module (PEM). This module is designed to bridge the gap between high-level semantic features and low-level structural cues. By fusing intermediate representations from the image encoder with the precise boundary priors extracted by the Edge-Aware Module (EAM), the PEM synthesizes boundary-informed initial masks. Unlike standard sparse prompts (points or boxes) that lack local geometry, these initial masks encapsulate both global context and fine-grained structural integrity. Through an adaptive refinement process supervised by ground-truth labels, the PEM refines these masks into mask-form prompts that are further encoded as dense prompt embeddings by the SAM prompt encoder. This ensures that the final mask prediction is guided by a rich, multi-modal understanding of the target anatomy, effectively resolving the boundary ambiguities common in ultrasound imaging.


Extensive experiments on six ultrasound datasets demonstrate that EP-SAM consistently outperforms state-of-the-art SAM-based methods, both without prompts and with single-point prompts, highlighting its strong capabilities.

\section{Related Work}

\subsection{Ultrasound Image Segmentation}

Ultrasound image segmentation is a fundamental task in computer-aided diagnosis, yet its real-time, reflection-based imaging mechanism often results in low image quality, weak contrast, and ambiguous boundaries, posing significant challenges for accurate segmentation.

Deep learning has substantially advanced ultrasound segmentation, starting from U-Net~\cite{ronneberger2015u} and extending to numerous variants that incorporate attention mechanisms~\cite{oktay2018attention}, multi-scale feature representations~\cite{zhou2018unet++}, and transformer-based global context modeling~\cite{chen2024transunet, cao2022swin, huang2022missformer}. While these architectures effectively capture hierarchical and contextual information, their performance is often limited by poor generalization across heterogeneous ultrasound domains and a reduced ability to handle boundary ambiguity and noise-induced artifacts.

Unlike task-specific architectures, we fine-tune the Segment Anything Model (SAM) to efficiently adapt it to ultrasound image segmentation by introducing an Edge-Aware Module and a Prompt Enhanced Module that strengthen coordination among its core components.

\subsection{Adapting SAM to Medical Ultrasound Images}

Although SAM~\cite{kirillov2023segment} achieves strong prompt-driven zero-shot segmentation on natural images, its performance degrades in ultrasound imaging due to pronounced domain shifts, weak boundaries, and scarce semantic cues~\cite{zhang2024segment}.

Existing efforts to adapt SAM for ultrasound image segmentation can be broadly categorized into three paradigms. 
(1) Mask decoder-level adaptation: MedSAM~\cite{ma2024segment} fine-tunes only the mask decoder to transfer SAM to medical domains. 
(2) Image encoder-level adaptation: SAMed~\cite{zhang2023customized} introduces LoRA-based fine-tuning within the SAM image encoder, while MSA~\cite{wu2025medical} inserts lightweight adapters to enhance domain-specific representation learning. 
(3) Prompt encoder-level adaptation: CC-SAM~\cite{gowda2024cc} proposes a medical large language model-guided prompting strategy, and PGP-SAM~\cite{yan2025pgp} designs a few-shot prompt generation scheme based on intra-class and inter-class prototypes.

Despite these advances, most existing methods adapt SAM by treating the image encoder, prompt encoder, and mask decoder as independent modules, thereby neglecting their intrinsic synergy and under-utilizing the rich structural and semantic information in intermediate image features. In contrast, our method exploits boundary cues embedded in the SAM image encoder’s intermediate representations to guide prompt enhancement, substantially strengthening the interaction between the image and prompt encoders.

\subsection{Edge-Aware Medical SAM}
Directly applying SAM to medical images often yields poor segmentation due to inherently weak and ambiguous boundaries. To address this, recent works have explored edge-aware enhancements. For example, SAM-EG~\cite{trinh2024sam} extracts edges using classical operators such as Sobel, Canny, and Laplacian, and fuses them with the encoder features. UniUltra~\cite{li2025uniultra} adapts SAM to ultrasound by inserting Sobel-equipped adapters into the image encoder. EASAM~\cite{zhang2025easam} adds a U-Net-like branch to capture structural cues, while BA-SAM~\cite{chen2024ba} derives boundary information via erosion and dilation in the feature space.

Unlike approaches that treat edge detection as an isolated auxiliary task or rely on external heuristics, our method extracts segmentation-relevant boundary cues directly from SAM’s intermediate encoder features. By integrating these cues into the prompt generation process, we eliminate the ``boundary-blind" limitation of standard adapters, enabling SAM to produce masks with significantly sharpened and anatomically precise boundaries.

\section{METHODOLOGY}

\begin{figure*}[t]
\centering
\includegraphics[width=0.9\textwidth]{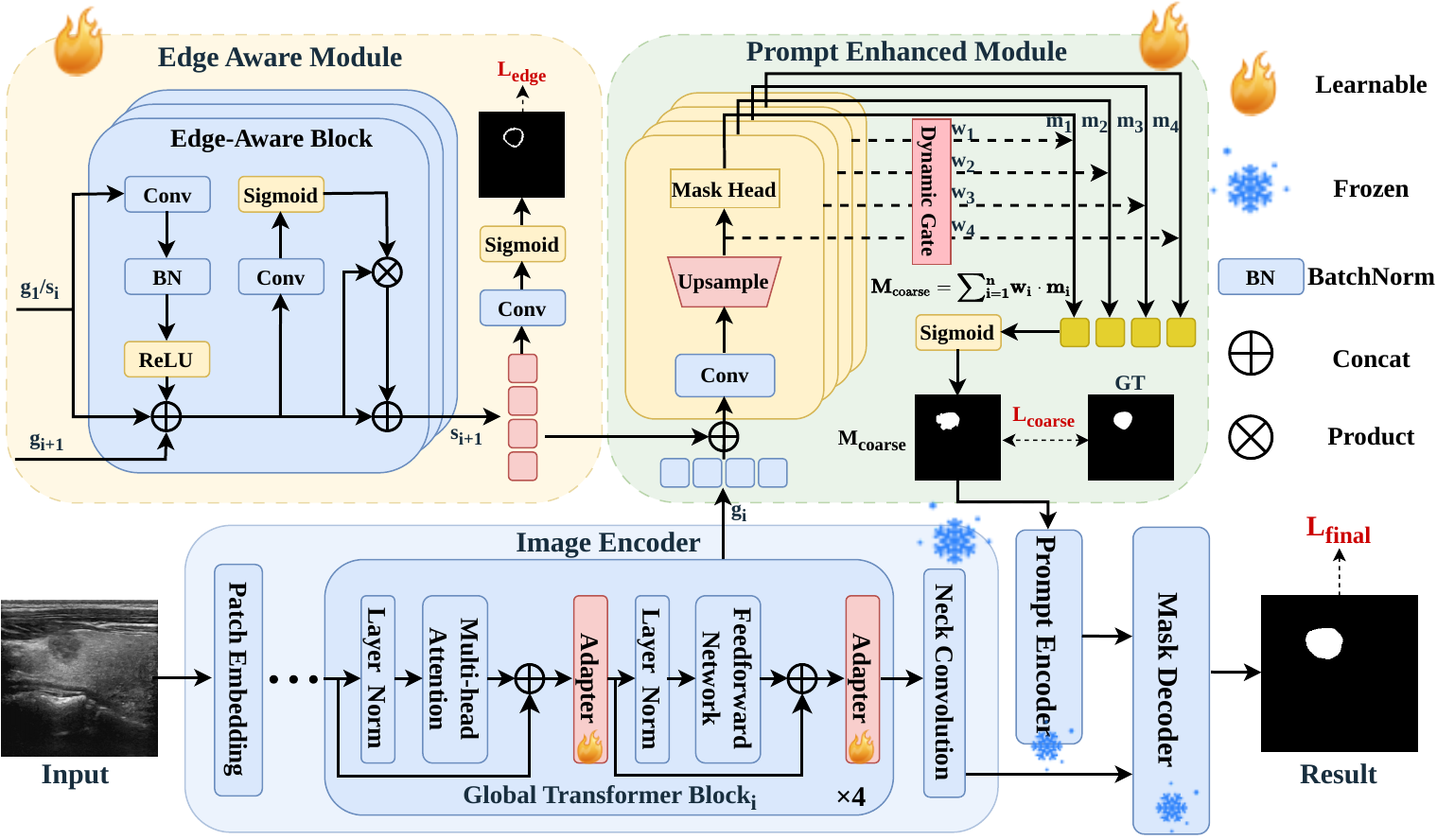} 
\caption{Overview of EP-SAM with two key modules: 
an edge-aware module that enhances contour modeling and a prompt enhanced module that injects coarse-to-fine mask priors for enhancing synergy of individual components.}
\label{fig2}
\end{figure*}

Given an input image $I \in \mathbb{R}^{H \times W \times 3}$, our goal is to predict an instance-level segmentation map $\mathbf{S} \in \mathbb{R}^{H \times W}$, where each pixel is assigned to a specific object instance. As illustrated in Figure~\ref{fig2}, our framework, EP-SAM, builds upon SAM and incorporates Mona Adapters~\cite{yin20255} into the image encoder for lightweight fine-tuning. EP-SAM leverages multi-level encoder features to enhance the synergy between SAM’s image encoder and prompt encoder, improving the model’s ability to produce consistent and precise instance masks.

\subsection{Edge-Aware Module}

To explicitly capture structural and boundary cues preserved in the early stages of SAM's image encoder, we introduce an \emph{Edge-Aware Module} (EAM). Unlike prior adapter-based tuning strategies that emphasize high-level semantic adaptation, EAM operates on intermediate transformer features, retaining fine-grained spatial structures critical for ultrasound instance segmentation.
As illustrated in Figure~\ref{fig2}, let $\mathbf{g}_i \in \mathbb{R}^{B \times C \times H/8 \times W/8}$ for $i = 1, 2, 3, 4$ denote feature maps extracted from Global Transformer Blocks of SAM's image encoder, where $B$ is  batch size and $C$ is the number of channels. 
The Edge-Aware Module (EAM) consists of a sequence of \emph{Edge-Aware Blocks} (EABs), which progressively refine edge-aware representations in a coarse-to-fine manner.

\paragraph{Edge-Aware Block}

Each Edge-Aware Block (EAB) performs residual convolutional refinement followed by gated spatial interaction. To maintain a unified coarse-to-fine workflow and avoid manual initialization, we do not define an initial edge feature $\mathbf{s}_1$; the first block directly refines $\mathbf{g}_1$ and combines it with $\mathbf{g}_2$ to produce $\mathbf{s}_2$, enabling meaningful boundary representations to be learned from transformer features.

Specifically, at stage $i$, the input feature (i.e., $\mathbf{g}_1$ for $i{=}1$ or $\mathbf{s}_i$ otherwise) is first refined by a residual convolutional refinement step to enhance local structural patterns. The refined feature is then concatenated with the subsequent transformer feature $\mathbf{g}_{i+1}$ and modulated by a gated spatial interaction, which adaptively emphasizes boundary-consistent responses while suppressing texture-dominated activations. The output of this block yields the next edge-aware representation $\mathbf{s}_{i+1}$. 

By progressively applying EABs across transformer stages, EAM incrementally strengthens boundary-aware representations under multi-level contextual guidance.

\paragraph{Edge Supervision and Prompt Guidance}
To explicitly enforce edge awareness, each edge-aware feature is projected to a single-channel edge probability map:
\begin{equation}
\mathbf{E}_i = \sigma\big( \mathcal{P}(\mathbf{s}_i) \big),
\end{equation}
where $\mathcal{P}(\cdot)$ denotes a lightweight projection head, and $\sigma(\cdot)$ is the element-wise sigmoid function that converts the projected feature into a probability map for loss computation against the boundary ground truth. All edge predictions are supervised with an edge-aware loss, encouraging consistent contour localization across stages.

To improve the interaction between SAM's image and prompt encoders, the multi-block features are forwarded to the \emph{Prompt Enhanced Module} (PEM), where edge-aware representations provide structural guidance to progressively generate high-quality prompts that steer coarse mask prediction.

\subsection{Prompt Enhanced Module}

We introduce the \textbf{Prompt Enhanced Module (PEM)} to further strengthen the synergy between SAM's image and prompt encoders. PEM integrates multi-block features from the image encoder with edge-aware outputs from the Edge-Aware Module (EAM) to produce coarse mask hypotheses, which serve as structured and semantic guidance for prompt encoding. Unlike approaches that rely on external models or handcrafted priors, PEM derives these coarse spatial prompts directly from SAM's multi-level transformer representations, ensuring self-contained and adaptive prompt generation.

Specifically, the intermediate features from the Global Transformer Blocks and their corresponding outputs from the Edge-Aware Blocks are concatenated, followed by a lightweight $1 \times 1$ convolution to reduce the channel dimension, simplifying the feature representation. The adapted features are then bilinearly upsampled to a unified resolution of $H/2 \times W/2$, matching the input size for prompt encoder.

PEM predicts a set of coarse mask hypotheses $\{ \mathbf{m}_1, \mathbf{m}_2, \mathbf{m}_3, \mathbf{m}_4 \}$ using parallel $1 \!\times\! 1$ convolutional heads. To adaptively fuse multi-level information, a dynamic gating mechanism computes level-wise weights $\mathbf{w} \in \mathbb{R}^n$ via global pooling followed by a lightweight gating network with softmax normalization, where $n$ is the number of selected stages. The final coarse mask is obtained as
\begin{equation}
\mathbf{M}_{\text{coarse}} = \sum_{i=1}^{n} w_i \cdot \mathbf{m}_i.
\end{equation}

The coarse mask is directly used as a mask prompt and fed into SAM's original prompt encoder, enabling prompt enhancement without modifying the mask decoder. In this way, PEM reliably transforms multi-level transformer features into explicit prompt-level guidance, thereby strengthening the collaborative interaction between image and prompt encoders.

\subsection{Loss Function}

We employ a unified multi-level supervision scheme to jointly optimize the Edge-Aware Module, the Prompt Enhanced Module, and the SAM mask decoder, with supervision applied to the predicted edge map, the intermediate coarse mask, and the final segmentation output.

All segmentation-related objectives are optimized using a hybrid loss that combines Dice loss and weighted binary cross-entropy (BCE), which effectively alleviates class imbalance and stabilizes training:
\begin{equation}
\mathcal{L}_{\text{seg}} = (1 - \alpha)\,\mathcal{L}_{\text{bce}}(\hat{Y}, Y) + \alpha\,\mathcal{L}_{\text{dice}}(\hat{Y}, Y),
\label{eq:seg_loss}
\end{equation}
where $\hat{Y}$ and $Y$ denote the predicted mask and the corresponding ground truth, respectively.

The Edge-Aware Module uses an edge loss $\mathcal{L}_{\text{edge}}$, the Prompt Enhanced Module uses a coarse mask loss $\mathcal{L}_{\text{coarse}}$, and the final SAM output is supervised by $\mathcal{L}_{\text{final}}$. The ground-truth boundary map for $\mathcal{L}_{\text{edge}}$ is directly extracted from the ground-truth segmentation mask via Canny edge detection. All three terms follow the formulation of $\mathcal{L}_{\text{seg}}$ in Eq.~\eqref{eq:seg_loss}, but are applied to different targets, namely the predicted edge map, the intermediate coarse mask, and the final segmentation output. The total objective is:
\begin{equation}
\mathcal{L}_{\text{total}} =
\mathcal{L}_{\text{edge}} +
\mathcal{L}_{\text{coarse}} +
\mathcal{L}_{\text{final}}.
\end{equation}
The weights are all set to 1 to equally balance the three supervision terms during optimization.

\section{Experiment}

\begin{table*}[t] 
\centering 
\caption{Quantitative results comparing our method with prior approaches on TN3K, BUSI, and CAMUS benchmarks. Metrics include Dice (\%) and Hausdorff Distance (HD). Bold indicates the best performance and underline denotes the second best.} 
\label{tab1} 
\small
\setlength{\tabcolsep}{3pt}
\renewcommand{\arraystretch}{1.0} 
\begin{tabularx}{\textwidth}{c*{11}{>{\centering\arraybackslash}X}} 
\toprule 
\multirow{2}{*}{\textbf{Method}} & 
\multirow{2}{*}{\textbf{Prompt}} &
\multicolumn{2}{c}{\textbf{TN3K}} & 
\multicolumn{2}{c}{\textbf{BUSI}} & 
\multicolumn{2}{c}{\textbf{CAMUS}} &
\multicolumn{2}{c}{\textbf{AVERAGE}}\\ 
\cmidrule(lr){3-4} \cmidrule(lr){5-6} \cmidrule(lr){7-8} \cmidrule(lr){9-10} 
& & Dice~($\uparrow$) & HD~($\downarrow$) & Dice~($\uparrow$) & HD~($\downarrow$) & Dice~($\uparrow$) & HD~($\downarrow$) & Dice~($\uparrow$) & HD~($\downarrow$) \\ 
\midrule 

U-Net~\cite{ronneberger2015u} & $\times$ & 79.01 & 34.12 & 78.11 & 33.60 & 86.86 & 16.87 & 81.33 & 28.20 \\ 
CE-Net~\cite{gu2019net} & $\times$ & 80.37 & 32.79 & 81.60 & 29.19 & 86.47 & 16.66 & 82.81 & 26.21 \\ 
CPFNet~\cite{feng2020cpfnet} & $\times$ & 79.43 & 33.07 & 80.56 & 27.98 & 86.68 & 16.51 & 82.22 & 25.85 \\ 
CA-Net~\cite{gu2020net} & $\times$ & 80.52 & 33.65 & 81.88 & 28.67 & 87.21 & 16.24 & 83.20 & 26.19 \\ 
SETR~\cite{zheng2021rethinking} & $\times$ & 67.80 & 44.11 & 68.22 & 40.37 & 86.20 & 18.27 & 74.07 & 34.25 \\ 
MISSFormer~\cite{huang2022missformer} & $\times$ & 79.42 & 32.85 & 78.43 & 33.10 & 86.57 & 16.50 & 81.47 & 27.48 \\ 
TransUNet~\cite{chen2024transunet} & $\times$ & 81.44 & 30.98 & 82.22 & 27.54 & 87.20 & 16.36 & 83.62 & 24.96 \\ 
TransFuse~\cite{zhang2021transfuse} & $\times$ & 78.50 & 32.44 & 73.52 & 34.95 & 86.77 & 17.25 & 79.60 & 28.21 \\
AAU-Net~\cite{chen2022aau} & $\times$ & 82.28 & 30.53 & 80.81 & 28.96 & 86.98 & 16.49 & 83.36 & 25.33 \\ 
SwinUNet~\cite{cao2022swin} & $\times$ & 70.08 & 44.13 & 67.23 & 47.02 & 84.46 & 20.25 & 73.92 & 37.13 \\ 
FAT-Net~\cite{wu2022fat} & $\times$ & 80.45 & 32.77 & 82.16 & 28.55 & 87.19 & \textbf{15.93} & 83.27 & 25.75 \\ 
H2Former~\cite{he2023h2former} & $\times$ & 82.48 & 30.58 & 81.48 & 27.84 & \underline{87.31} & 16.60 & 83.76 & 25.01 \\ 
SAMUS~\cite{lin2024beyond} & $\times$ & 81.41 & 31.44 & 84.33 & 26.37 & 86.74 & 16.85 & 84.16 & 24.89\\
MSA~\cite{wu2025medical} & $\times$ & 73.68 & 31.11 & 78.34 & 27.25 & 86.33 & 16.91 & 79.45 & 25.09 \\
UN-SAM~\cite{chen2025sam} & $\times$ & \underline{82.88} & \underline{30.04} & \underline{84.76} & \underline{25.19} & 87.03 & 16.51 & \underline{84.89} & \underline{23.91} \\
Ours & $\times$ & \textbf{83.65} & \textbf{28.65} & \textbf{85.39} & \textbf{23.90} & \textbf{87.47} & \underline{16.02} & \textbf{85.50} & \textbf{22.86} \\
\midrule

SAMUS~\cite{lin2024beyond} & point & \underline{84.45} & \underline{28.22} & \underline{85.77} & \underline{25.49} & \underline{87.46} & 16.74 & \underline{85.89} & \underline{23.48}\\ 
MSA~\cite{wu2025medical} & point & 75.68 & 28.95 & 81.89 & 24.10 & 87.10 & 16.81 & 81.56 & 23.29   \\
Ours & point & \textbf{85.48} & \textbf{27.26} &\textbf{87.70} & \textbf{22.63}& \textbf{87.66} &  \textbf{16.65} & \textbf{86.95} & \textbf{22.18} \\ 
\midrule
\rowcolor{gray!15}
CC-SAM~\cite{gowda2024cc} & LLM & 85.20 & 27.10 & 87.01 & 24.22 & 88.25 & 16.11 & 86.82 & 22.48 \\ 

\bottomrule 
\end{tabularx} 
\end{table*}

\subsection{Experimental Settings}
\subsubsection{Datasets}
We evaluate our method on three widely-used ultrasound segmentation benchmarks covering different anatomical structures, including breast lesions, thyroid nodules, and myocardium. For breast ultrasound segmentation, we employ BUSI~\cite{al2019deep} and UDIAT ~\cite{byra2020breast}. The BUSI dataset is randomly divided into training, validation, and test sets with a ratio of 7:1:2. For thyroid nodule segmentation, we use the TN3K~\cite{gong2023thyroid} and DDTI~\cite{pedraza2015open} datasets. Following the experimental protocol of TRFE~\cite{gong2023thyroid}, TN3K is split into training, validation and testing subsets.
For myocardium segmentation, we adopt the CAMUS~\cite{leclerc2019deep} and HMC-QU~\cite{degerli2024early} datasets. 
CAMUS is divided into a train set and a test set first according to the challenge~\cite{leclerc2019deep}. Then, we randomly select 10\% of patients from the train set to validate the model and the rest data as the final training data.

\subsubsection{Evaluation Protocol}

We conduct in-domain and cross-domain evaluations to assess segmentation accuracy and generalization. Models are trained and evaluated on BUSI, TN3K, and the myocardium (MYO) category of CAMUS under in-domain settings. To examine whether the proposed method preserves SAM’s generalization capability, we directly test the trained models on unseen datasets, including UDIAT, DDTI, and HMC-QU, without fine-tuning. All methods follow the same evaluation protocol and are trained and evaluated under two prompt settings: (i) no prompts and (ii) sparse point prompts randomly sampled from ground truth masks. Performance is measured using the Dice coefficient (Dice) and the maximum Hausdorff Distance (HD).

\subsubsection{Implementation Details}

All experiments are implemented in PyTorch, following the hyperparameter settings and position adapter configurations of SAMUS~\cite{lin2024beyond}. All ultrasound images are resized to $256 \times 256$ before being fed into the network. We adopt SAM-B as the backbone throughout all experiments. For the image encoder, Mona adapters~\cite{yin20255} are inserted into each transformer block and fine-tuned during training, while the remaining parameters of the image encoder backbone are kept frozen. The prompt encoder and mask decoder of SAM are also frozen during training. The model is optimized using Adam with an initial learning rate of $1\times10^{-4}$ and a batch size of 8 for 400 epochs. Following SAMUS~\cite{lin2024beyond}, $\alpha$ is fixed to $0.8$ throughout all experiments. All experiments are conducted on a single NVIDIA RTX 4090 GPU (24 GB). 

\subsection{Comparison with SOTA Methods}

\begin{figure}[t]
\centering
\includegraphics[width=0.95\columnwidth]{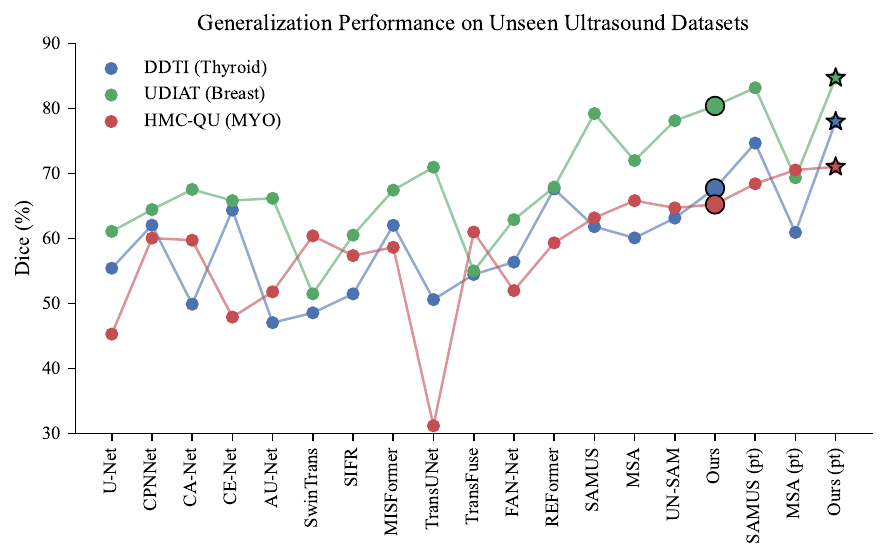} 
\caption{Generalization performance (Dice \%) on three unseen ultrasound datasets (DDTI~\cite{pedraza2015open}, UDIAT~\cite{byra2020breast}, HMC-QU~\cite{degerli2024early}).}
\label{fig3}
\end{figure}

\noindent\textbf{Quantitative comparisons} are reported in Table~\ref{tab1}.
EP-SAM is evaluated on TN3K, BUSI, and CAMUS datasets, comparing its performance against representative CNN-, Transformer-, and SAM-based methods, including the self-prompting method UN-SAM~\cite{chen2025sam}. \textbf{Without prompts}, EP-SAM achieves the best performance on TN3K and BUSI, and attains the highest Dice and the second-best HD on CAMUS, yielding the highest average Dice of 85.50\% and the lowest average HD of 22.86, outperforming strong non-prompt baselines such as UN-SAM and SAMUS.
These improvements are particularly pronounced on BUSI and TN3K, where boundary ambiguity and low contrast are severe, demonstrating that our EAM and PEM effectively leverage intermediate encoder features and enhance the coordination between SAM’s image encoder and prompt encoder.
\textbf{With point prompts}, EP-SAM further improves accuracy and consistently ranks first across all datasets, achieving an average Dice of 86.95\% and an average HD of 22.18, surpassing prompt-based methods including SAMUS and MSA. Compared with LLM-assisted approaches such as CC-SAM, EP-SAM achieves comparable or superior performance using only single-point prompts, highlighting that explicitly modeling feature-level interactions between SAM components can yield high segmentation accuracy with minimal prompt complexity.

Moreover, EP-SAM attains the best Dice on three unseen datasets (DDTI, UDIAT, and HMC-QU) under both prompt-free and single-point prompt settings (Figure~\ref{fig3}), demonstrating strong cross-dataset generalization, which further indicates that enhancing the synergy between SAM modules preserves the rich semantic and structural information across domains.

\begin{figure}[t]
\centering
\includegraphics[width=0.95\columnwidth]{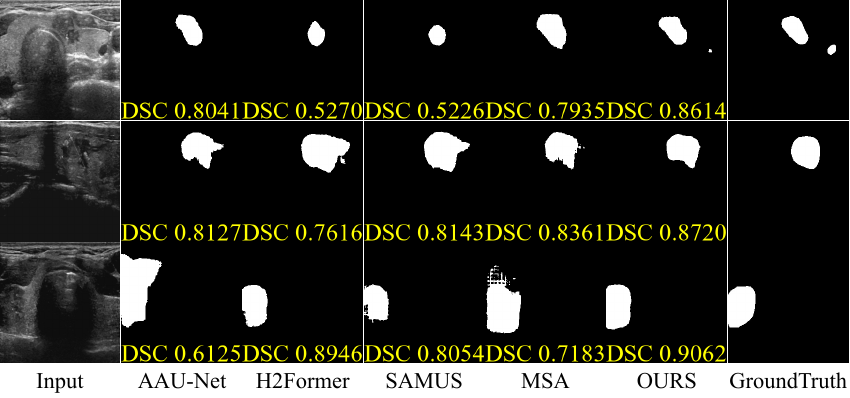}
\caption{Visualization of EP-SAM and other SOTA methods.}
\label{fig4}
\end{figure}

\noindent\textbf{Qualitative Analysis.}
Figure~\ref{fig4} compares the segmentation results produced by different methods. Ultrasound image segmentation is particularly challenging due to low contrast, heterogeneous textures, and ambiguous object boundaries. Despite these difficulties, our method yields more accurate and stable predictions, especially along object boundaries and in regions with weak structural cues, demonstrating its robustness in challenging ultrasound scenarios.

\begin{table}[t]
\centering
\caption{Ablation of individual components.}
\label{tab2}
\begin{tabularx}{\columnwidth}{
  >{\centering\arraybackslash}X
  >{\centering\arraybackslash}X
  >{\centering\arraybackslash}X
  >{\centering\arraybackslash}X
  >{\centering\arraybackslash}X
  >{\centering\arraybackslash}X}
\toprule
\multicolumn{2}{c}{\textbf{Components}} &
\multicolumn{2}{c}{\textbf{TN3K}} &
\multicolumn{2}{c}{\textbf{BUSI}} \\
\cmidrule(lr){1-2}\cmidrule(lr){3-4}\cmidrule(lr){5-6}
EAM & PEM & Dice$\uparrow$ & HD$\downarrow$ & Dice$\uparrow$ & HD$\downarrow$ \\
\midrule
$\times$ & $\times$ & 81.18 & 31.18 & 81.61 & 29.05 \\
$\checkmark$ & $\times$ & 82.40 & 30.65 & 82.75 & 28.48 \\
$\times$ & $\checkmark$ & 83.10 & 28.92 & 84.84 & 25.26 \\
$\checkmark$ & $\checkmark$ & \textbf{83.65} & \textbf{28.65} & \textbf{85.39} & \textbf{23.90} \\
\bottomrule
\end{tabularx}
\end{table}

\begin{table}[t]
\centering
\caption{The effect of the transformer block in PEM.}
\label{tab3}
\begin{tabularx}{\columnwidth}{
  >{\centering\arraybackslash}X
  >{\centering\arraybackslash}X
  >{\centering\arraybackslash}X
  >{\centering\arraybackslash}X
  >{\centering\arraybackslash}X
  >{\centering\arraybackslash}X}
\toprule
\multicolumn{4}{c}{\textbf{Block Features}} &
\multicolumn{2}{c}{\textbf{BUSI}} \\
\cmidrule(lr){1-4}\cmidrule(lr){5-6}
$ \mathbf{g}_1 $ & $ \mathbf{g}_2 $ & $ \mathbf{g}_3 $ & $ \mathbf{g}_4 $ & Dice$\uparrow$ & HD$\downarrow$ \\
\midrule
$\checkmark$ & $\times$ & $\times$ & $\times$ & 83.85 & 27.57 \\
$\times$ & $\checkmark$ & $\times$ & $\times$ & 84.71 & 26.11 \\
$\times$ & $\times$ & $\checkmark$ & $\times$ & 84.52 & 26.98 \\
$\times$ & $\times$ & $\times$ & $\checkmark$ & 84.88 & 25.68 \\
$\checkmark$ & $\checkmark$ & $\checkmark$ & $\checkmark$ & \textbf{85.39} & \textbf{23.90} \\
\bottomrule
\end{tabularx}
\end{table}

\subsection{Ablation Study}
Table~\ref{tab2} presents an ablation study analyzing the contribution of each component in EP-SAM on the TN3K and BUSI datasets. The baseline model without EAM or PEM serves as a reference. Introducing the Edge-Aware Module (EAM) improves the Dice score from 81.18\% to 82.40\% on TN3K and from 81.61\% to 82.75\% on BUSI, accompanied by consistent reductions in HD, indicating the benefit of explicit boundary modeling. Incorporating the Prompt Enhanced Module (PEM) leads to more pronounced gains, yielding a Dice improvement of +1.92\% on TN3K and +3.23\% on BUSI over the baseline, highlighting the effectiveness of self-generated prompt cues. When both EAM and PEM are jointly applied, EP-SAM achieves the best performance across all settings, reaching Dice scores of 83.65\% and 85.39\% on TN3K and BUSI, respectively. This improvement arises from the enhanced synergy between SAM’s image encoder and the Prompt Encoder facilitated by EAM and PEM. These results demonstrate the complementary nature of edge-aware feature extraction and prompt enhancement for robust ultrasound segmentation. Moreover, we also validate Mona adapters~\cite{yin20255}: on BUSI, they achieve 81.61\% Dice versus 80.23\% with linear adapters, confirming their superior feature modulation. Additional details are provided in the supplementary material.

Table~\ref{tab3} presents an ablation study on the contribution of individual transformer block features within the Prompt Enhanced Module. Using a single block, $ \mathbf{g}_4 $ achieves the highest Dice (84.88\%) and lowest HD (25.68), suggesting that higher-level features provide stronger semantic guidance. Lower-level features ($ \mathbf{g}_1 $--$ \mathbf{g}_3 $) also contribute, with $ \mathbf{g}_2 $ and $ \mathbf{g}_3 $ yielding moderate gains. Combining all four blocks consistently improves performance, reaching 85.39\% Dice and 23.90 HD, highlighting the complementary effect of multi-block features for generating robust, structure-aware prompts.

\section{Conclusion}

We present EP-SAM, a novel framework for ultrasound segmentation that enhances collaboration between SAM's image and prompt encoders. By integrating edge-aware module and prompt enhanced module, EP-SAM refines boundary structures and enriches prompt embeddings. Extensive evaluations on ultrasound benchmarks show that EP-SAM outperforms state-of-the-art CNN-, Transformer-, and SAM-based methods in both prompt-free and point-prompt settings, achieving superior accuracy, boundary precision, and generalization. These results emphasize the benefit of improving the synergy between SAM’s components. Future work will focus on refining self-supervised prompts, integrating multi-scale edges, and enabling lightweight deployment for clinical use.

\section*{Acknowledgment}
This work was supported by National Key Research and Development Program of China (Grant No. 2023YFC2705701), the New Cornerstone Science Foundation through the XPLORER PRIZE, National Science Foundation for Distinguished Young Scholars of China (Grant No. 62225113), WHU-Kingsoft Joint Lab and the super-computing system in the Supercomputing Center of Wuhan University.

\bibliographystyle{IEEEtranS}
\bibliography{icme2026references}

\end{document}